\documentclass[conference]{IEEEtran}
\IEEEoverridecommandlockouts
\def\BibTeX{{\rm B\kern-.05em{\sc i\kern-.025em b}\kern-.08em
    T\kern-.1667em\lower.7ex\hbox{E}\kern-.125emX}}
\usepackage{cite}
\usepackage{amsmath,amssymb,amsfonts}
\usepackage{graphicx}
\usepackage{textcomp}
\usepackage{xcolor}
\usepackage{booktabs}
\usepackage{hyperref}
\usepackage{multirow}
\usepackage{url}
\usepackage{algorithm}
\usepackage{algpseudocode}
\usepackage{amsmath,amsfonts,bm}

\begin{document}

\title{Blockwise Principal Component Analysis for monotone missing data imputation and dimensionality reduction}


\author{\IEEEauthorblockN{Tu T. Do, Mai Anh Vu, Tuan L. Vo}
 \IEEEauthorblockA{
\textit{Dept. of Mathematics and Computer Science} \\
 \textit{University Of Science}\\
 \textit{Vietnam National University in Ho Chi Minh City}\\
 Ho Chi Minh city, Vietnam}
\and
\and
\IEEEauthorblockN{Hoang Thien Ly}
 \IEEEauthorblockA{
\textit{ Faculty of Mathematics and Information Science  } \\
 \textit{Warsaw University of Technology}\\
Warsaw, Poland}
\and
\IEEEauthorblockN{Thu Nguyen, Steven Hicks, Pål Halvorsen, Michael A. Riegler}
\IEEEauthorblockA{\textit{Department of Holistic Systems} \\
\textit{Simula Metropolitan}\\
Oslo, Norway}
\and
\IEEEauthorblockN{Binh T. Nguyen}
 \IEEEauthorblockA{\textit{AISIA Research Lab} \\
\textit{Department of Computer Science} \\
 \textit{University Of Science}\\
 \textit{Vietnam National University in Ho Chi Minh City}\\
 Ho Chi Minh city, Vietnam}
}

\maketitle

\begin{abstract} 
Monotone missing data is a common problem in data analysis. However, imputation combined with dimensionality reduction can be computationally expensive, especially with the increasing size of datasets. To address this issue, we propose a Blockwise Principal Component Analysis Imputation (BPI) framework for dimensionality reduction and imputation of monotone missing data. The framework conducts Principal Component Analysis on the observed part of each monotone block of the data and then imputes on merging the obtained principal components using a chosen imputation technique. BPI can work with various imputation techniques and can significantly reduce imputation time compared to conducting dimensionality reduction after imputation. This makes it a practical and efficient approach for large datasets with monotone missing data. Our experiments validate the improvement in speed while achieving an accuracy that is comparable to the common strategy of imputation prior to dimensional reduction.  
\end{abstract}


\section{Introduction}


Dimensionality reduction is a technique used in machine learning and data analysis to reduce the number of features in a dataset while preserving the essential information. It is an essential technique 
for several reasons, including simplifying the dataset's structure, reducing computational complexity, ameliorating overfitting issues, removing noise and redundancy in data, and enabling visualization because it helps transform the data into a lower-dimensional space to visualize and interpret. 
In addition, dimensionality reduction can improve the performance of machine learning models by reducing overfitting and increasing generalization ability.

However, if missing values exist in the dataset, it can be challenging to perform dimensionality reduction because many techniques require a complete dataset. In addition, monotone missing data (i.e., data where if a certain data point is missing, then all subsequent data points in that sequence or series are also missing) is a common problem in practice \cite{NGUYEN20211,little2019statistical}. 
For example, in the following, $\mathcal{D}_1, \mathcal{D}_2$ are examples of datasets with monotone missing data patterns, while $\mathcal{D}_3$ is not.
\begin{align*}
\small
\mathcal{D}_1 &= \begin{pmatrix}
    2 & 3 & 5 & 7 & 9 \\
    1 & 2 & 4 & * & *\\
    3& 2& 6 & * & *    
\end{pmatrix}, \\ \\
\mathcal{D}_2 &= \begin{pmatrix}
    8 & 3 & 5 & 7 & 1 \\
    1 & 2 & 4 & * & *\\
    3& 2& * & * & *     
\end{pmatrix}, \\ \\
\mathcal{D}_3 &= \begin{pmatrix}
    8 & 3 & 5 & 7 & 1 \\
    1 & 2 & 4 & * & *\\
    3& 2& * & 1 & 12     
\end{pmatrix}.
\end{align*}
The above is just a simple toy dataset for illustration. In practice, such a monotone pattern can happen in various scenarios. For example, in longitudinal studies \cite{hedeker2006longitudinal}, participants may miss follow-up assessments, which leads to the missingness of subsequent time points. Similarly, in various clinical trials where patients are monitored for a particular health outcome over a period of time, if a patient drops out of the study after certain visits, all subsequent data points for that patient are missing. This also creates a monotone missing pattern.

It is common to deal with missing data by filling in the missing entries via some imputation technique. Various imputation techniques \cite{vu2023conditional,stekhoven2012missforest,buuren2010mice} have been developed to deal with different scenarios and different types of data. However, many widely used imputation techniques are not suitable for large datasets. This is illustrated via experiments by Nguyen et al.~\cite{NGUYEN2022108082}, where MICE \cite{buuren2010mice} and MissForest \cite{stekhoven2012missforest} are not evan able to finish imputation within three hours.  
Recently, PCAI \cite{nguyen2022principle} was introduced as a framework to speed up the imputation when there are many fully observed features in the data. Specifically, PCAI partitions the data into the fully observed features partition and the partition of features with missing data. After that, the imputation of the missing part is performed based on the union of the principal components of the fully observed and the missing part. 
However, even if imputation can be sped up, imputation and then conducting dimensionality reduction is still a computationally expensive approach. Hence, while it provides principle components directly from missing data, it is not a scalable approach.

In this work, we propose a block-wise principal component analysis Imputation (BPI) framework for dimensionality reduction for monotone missing data. BPI starts by conducting Principal Component Analysis (PCA) on the observed part of each block of the data and then imputes the data on the merge of the projection using some imputation technique. Since PCA is conducted on parts of the data before imputation, BPI significantly reduces the running time compared to the conventional strategy of imputing before reducing the dimension.

The main contributions are: (i) We introduce BPI, a novel framework for dimensionality reduction and imputation of missing data; (ii) We illustrate via experiments that BPI can work with various imputation methods and improve the running time significantly compared to conducting dimensionality reduction after imputation; (iii) 
We point out the drawbacks of our work and directions for future work.   

The paper is structured as follows: 
Section \ref{sec-related} summarizes some related works in the field. Next, in Section \ref{methodolody}, we introduce our proposed approach, Blockwise PCA Imputation (BPI). 
Next, in Section \ref{experiment}, we present experimental results to demonstrate our approach's effectiveness and discuss the implications of these results. Finally, we conclude the paper and outline potential future research directions in Section \ref{conclusion}.

\section{Related Works}\label{sec-related}
Most works addressing the issue of missing data predominantly focus on imputation strategies that aim to substitute the absent entries with plausible values, ensuring that the data becomes complete before subsequent analyses. Initial traditional methodologies included mean, mode, and median imputation. These are simplistic approaches wherein the central tendency of the observed values is utilized to replace the missing ones. Another early technique was regression-based imputation, where missing values are predicted based on relationships with other variables in a regression-like manner.

With the advance of computers and machine learning, more advanced techniques were formulated. One such method is the k-nearest neighbors imputation (KNNI), which considers $k$ similar instances (neighbors) from the dataset to compute a (weighted) average/majority voting as the imputed value. Additionally, decision tree-based techniques like missForest \cite{stekhoven2012missforest}, DMI algorithm \cite{rahman2013missing}, and DIFC algorithm \cite{nikfalazar2020missing} were introduced. These strategies leverage the hierarchical structure of decision trees to handle missing data efficiently. Moreover, matrix decomposition techniques such as Polynomial Matrix Completion \cite{fan2020polynomial} and SOFT-IMPUTE \cite{mazumder2010spectral} also stand out. They use matrix factorization to exploit the inherent low-rank structure of the data and thereby infer the missing entries.

Next, Bayesian and multiple imputation methods offers promising results in missing data scenarios. Bayesian network imputation \cite{hruschka2007bayesian} leverages the probabilistic relationships between variables, while methods like multiple imputations using Deep Denoising Autoencoders \cite{gondara2017multiple} employ neural networks to learn complex patterns in the data and impute accordingly. Additionally, Bayesian principal component analysis-based imputation \cite{audigier2016multiple} combines the power of PCA with Bayesian inference to handle missing data.

Over the recent years, the rise of deep learning techniques has brought about a significant shift in how this challenge is approached \cite{nelwamondo2013dynamic,choudhury2019imputation,gondara2017multiple,garg2018dl,leke2016missing,mohan2021graphical}.  These techniques, leveraging intricate neural network architectures, have the capability to model complex patterns and relationships in data, making them particularly effective for imputation tasks. However, one of the trade-offs of employing deep learning is its insatiable appetite for data. In contrast to traditional statistical imputation techniques, deep learning models often require vast amounts of data to train effectively and avoid overfitting.

In recent years, many hybrid imputation techniques have also been developed. For example, HPM-MI \cite{purwar2015hybrid} is a technique that uses K-means clustering to analyze various imputation techniques and apply the best one to a dataset. Another typical work is SvrFcmGa \cite{aydilek2013hybrid}, where a fuzzy c-means clustering hybrid approach is combined with support vector regression and a genetic algorithm.


Moreover, some studies have also concentrated on imputation for monotone missing data. For example, \cite{kombo2017multiple} compares the performance of fully conditional
specification imputation and multivariate normal imputation for monotone missing data with ordinal outcomes. 

However, the scalability of imputation remains a problem that requires more work to deal with various data types and missing patterns \cite{vu2023conditional}. Some works try to speed up the imputation process. For example, the PCAI framework \cite{nguyen2022principle} helps speed up an imputation algorithm by applying principal component analysis (PCA) on the features that have no missing entries. Next, the features with missing entries are imputed based on the union of itself with the principal components of the fully observed features. Their experiments show great performance in speed while maintaining competitive results compared to directly applying an imputation method. Later, the performance of the method for logistic regression was studied throughout in \cite{nguyen2023principal}. However, a limitation of PCAI is the margin of imputation speed gain is limited if the number of fully observed features is small.

Moreover, for large data sets, dimension techniques under missing data can be of great interest. In fact, there are some works on directly getting the principle components in PCA from missing data. 
For example, as discussed in \cite{folch2015pca}, nonlinear iterative partial least squares algorithm (NIPALS), which performs iterative regressions with observed data  \cite{folch2015pca}, or iterative imputation (IA) use iterative PCA models using predictions from previous models until convergence. As introduced in \cite{folch2015pca}, several new methods for building (PCA) models that handle missing data by using IA adaption. Among those,  IA combined with trimmed scores regression (TSR) combines PCA with multiple linear regression by trimming extreme values and regressing the remaining scores, showed promising performance. However, NIPALS may struggle with convergence if many values are missing, and IA may be time-consuming or require multiple iterations. However, the scalability of these methods has not been investigated throughout.

\section{Methodology}\label{methodolody}
When missing data is present in a dataset, it is a common practice to impute the data first before consequent processing and analysis. However, when a dataset is big, and dimension reduction is of interest, but the data contains missing values, it can be computationally expensive to impute the data as well. For monotone missing data, this burden can be ameliorated by applying PCA to the observed parts of each monotone block and imputing the union of the principle components obtained from each block. This is the basic idea of the BPI framework, which will now be detailed in this section. To start, assume that there is a dataset input $\boldsymbol{x}$ of $p$ features with the following monotone pattern:
\begin{align*}
\boldsymbol{x}=\begin{pmatrix} \boldsymbol{x}_{11}&\dots	 &\boldsymbol{x}_{1n_{k}}&\dots	 &\boldsymbol{x}_{1n_3}&...&\boldsymbol{x}_{1n_2} &...&\boldsymbol{x}_{1n_1}\\
\boldsymbol{x}_{21} &\dots&\boldsymbol{x}_{2n_{k}}&\dots&\boldsymbol{x}_{2n_3}&...&\boldsymbol{x}_{2n_2}&...&*\\
\boldsymbol{x}_{31} &\dots&\boldsymbol{x}_{3n_{k}}&\dots&\boldsymbol{x}_{3n_3}&...&*&...&*\\
\vdots &\ddots& \vdots&\ddots&\vdots&\ddots&\vdots&\ddots&\vdots\\
\boldsymbol{x}_{k1}&\dots &\boldsymbol{x}_{kn_{k}}&\dots&*&...&*&...&*
\end{pmatrix}.
\end{align*}
where $\boldsymbol{x}_{in_j}\in \mathbb{R}^{ p_j\times 1}$ and each column represents an observation. That is, there are $n_1$ observations available on the first $p_1$ variables, $n_2$ observations available {on the first $p_1+p_2$ variables}, and so on. Then the dimension of $\boldsymbol{x}$ is $\sum_{j=1}^kp_j$
 and one can partition the data into:
\begin{eqnarray}
\boldsymbol{x}_{1}&=&\begin{pmatrix}
\boldsymbol{x}_{11}& \dots &\boldsymbol{x}_{1n_k} & \hdots &\boldsymbol{x}_{1n_2}& \dots & \boldsymbol{x}_{1n_1}
\end{pmatrix}, \nonumber \\
\boldsymbol{x}_{2} &=&\begin{pmatrix}
\boldsymbol{x}_{21}&\dots & \boldsymbol{x}_{2n_k} &\dots& \boldsymbol{x}_{2n_2}
\end{pmatrix},  \nonumber \\
\vdots  \nonumber \\
\boldsymbol{x}_{k}&=&\begin{pmatrix}
\boldsymbol{x}_{k1}&\dots&\boldsymbol{x}_{kn_k}  \nonumber
\end{pmatrix},
\end{eqnarray}

where $\boldsymbol{x}_{1}$ has the size ${p_1\times n_1 }$, 
\dots, $\boldsymbol{x}_{k}$ has the size $p_k\times n_k$. 

Suppose that PCA upon $\boldsymbol{x}_i$ gives 
\begin{eqnarray}
\boldsymbol{z}_{1}&=&\begin{pmatrix}
\boldsymbol{z}_{11}& \dots &\boldsymbol{z}_{1n_k} & \hdots &\boldsymbol{z}_{1n_2}& \dots & \boldsymbol{z}_{1n_1}
\end{pmatrix}, \nonumber \\
\boldsymbol{z}_{2} &=&\begin{pmatrix}
\boldsymbol{z}_{21}&\dots & \boldsymbol{z}_{2n_k} &\dots& \boldsymbol{z}_{2n_2}
\end{pmatrix},  \nonumber \\
\vdots  \nonumber \\
\boldsymbol{z}_{k}&=&\begin{pmatrix}
\boldsymbol{z}_{k1}&\dots&\boldsymbol{z}_{kn_k}  \nonumber
\end{pmatrix}.
\end{eqnarray}
Here, $\boldsymbol{z}_i$ of size $q_i\times n_i, i =1,...,k$. 

Then, we can stack $\boldsymbol{z}_i$ together, and insert empty entries (here, we denote each empty entry by $*$) to present missing values. This gives 
\begin{equation}
\boldsymbol{z}^*=\begin{pmatrix}
\boldsymbol{z}_{11}& \dots &\boldsymbol{z}_{1n_k} & \hdots &\boldsymbol{z}_{1n_2}& \dots & \boldsymbol{z}_{1n_1}\\
\boldsymbol{z}_{21}&\dots & \boldsymbol{z}_{2n_k} &\dots& \boldsymbol{z}_{2n_2} & \dots & *\\
\vdots& \ddots &\vdots & \ddots &\vdots& \ddots & \vdots\\
\boldsymbol{z}_{k1}&\dots&\boldsymbol{z}_{kn_k}   & \hdots &*& \dots & *
\end{pmatrix}  
\end{equation}


Then, we can conduct imputation on $\boldsymbol{z}^*$ to get an imputed version $\boldsymbol{z}$ and consider it as an imputed reduced dimension version of $\boldsymbol{x}$. Since the imputation is conducted after stacking the observed reduced blocks. Therefore, there are fewer values to be imputed. 

Note that not all the blocks are of large dimensions. For example, it is possible that while $\boldsymbol{x}\in\mathbb{R}^{1000} $ but $\boldsymbol{x}_2\in \mathbb{R} ^2$. In such a case, it may be preferable not to reduce the dimension of $\boldsymbol{x}_2$.  
\begin{algorithm}
\caption{\textbf{BPI algorithm} }\label{alg-bpi}
\hspace*{\algorithmicindent} \textbf{Input:} 
\begin{enumerate}
    \item partitions $\boldsymbol{x}_1,...,\boldsymbol{x}_k$ of the training input $\boldsymbol{x}$ where $\boldsymbol{x}_i \in \boldsymbol{R}^{p_i\times n_i},$
    \item imputation algorithm $I$.
\end{enumerate}

\hspace*{\algorithmicindent} \textbf{Procedure:} 
\begin{algorithmic}[1]
    \For{ $i=1,...,k:$}
    
        Apply $PCA$ on $\mathbf{x}_i$ gives $\boldsymbol{z}_i = (\boldsymbol{z}_{i1},...,\boldsymbol{z}_{in_i}) \in \mathbb{R}^{q_i\times n_i}$
        \EndFor
    \State \begin{equation*}
\boldsymbol{z}^*=\begin{pmatrix}
\boldsymbol{z}_{11}& \dots &\boldsymbol{z}_{1n_k} & \hdots &\boldsymbol{z}_{1n_2}& \dots & \boldsymbol{z}_{1n_1}\\
\boldsymbol{z}_{21}&\dots & \boldsymbol{z}_{2n_k} &\dots& \boldsymbol{z}_{2n_2} & \dots & *\\
\vdots& \ddots &\vdots & \ddots &\vdots& \ddots & \vdots\\
\boldsymbol{z}_{k1}&\dots&\boldsymbol{z}_{kn_k}   & \hdots &*& \dots & *
\end{pmatrix}  
\end{equation*}
    \State Impute $\boldsymbol{z}^*$ by $I$ gives $\boldsymbol{z}$
\end{algorithmic}
\textbf{Return:} $\boldsymbol{z}$ as the imputed dimensional reduced version of $\boldsymbol{x}$.
\end{algorithm}

With the above reasoning, the algorithm is formalized in Algorithm \ref{alg-bpi}.

\paragraph{Example} As a toy example, suppose that we have a dataset where the input is
\begin{align*}
\boldsymbol{x}=\begin{pmatrix} 1 & 5& 2& 9 & 7 & 0 & 8\\
2 & 3 & 6 & 4 & 0 & 1 & 9 \\
3 & 1 & 8 & 3 & 5 & 2 & 0 \\
3 & 1 & 2 & 0 & 0 & * & * \\
0 & 4 & 1 & 3 & 2 & * & * \\
4 & 8 & 6 & * & * & * & * \\
9 & 1 & 2 & * & * & * & *
\end{pmatrix}.
\end{align*}
Then one can partition the $\boldsymbol{x}$ into:
\begin{eqnarray}
\boldsymbol{x}_{1}&=&\begin{pmatrix}
1 & 5& 2& 9 & 7 & 0 & 8\\
2 & 3 & 6 & 4 & 0 & 1 & 9 \\
3 & 1 & 8 & 3 & 5 & 2 & 0 
\end{pmatrix}, \nonumber \\
\boldsymbol{x}_{2} &=&\begin{pmatrix}
3 & 1 & 2 & 0 & 0\\
0 & 4 & 1 & 3 & 2
\end{pmatrix},  \nonumber \\
\boldsymbol{x}_{3}&=&\begin{pmatrix}
4 & 8 & 6\\
9& 1 & 2  \nonumber
\end{pmatrix}.
\end{eqnarray}
Suppose that PCA upon $\boldsymbol{x}_i$ gives $\boldsymbol{z}_i$ as follows
\begin{eqnarray}
\boldsymbol{z}_{1}&=&\begin{pmatrix}
0,5 & 2 & 1 & 0 & 1 & 0.9 & 2\\
1 & 0.7 & 0.3 & 2 & 0.5 & 1 & 1
\end{pmatrix}, \nonumber \\
\boldsymbol{z}_{2} &=&\begin{pmatrix}
1& 3 & 0.7 &0 & 0.3
\end{pmatrix},  \nonumber \\
\boldsymbol{z}_{3}&=&\begin{pmatrix}
2&0.5&1  \nonumber
\end{pmatrix}.
\end{eqnarray}

Then, we can stack $\boldsymbol{z}_i$ together and insert empty entries (here, we denote each empty entry by $*$) to present missing values. This gives 
\begin{equation}
    \boldsymbol{z}^*=\begin{pmatrix}
                    0.5 & 2 & 1 & 0 & 1 & 0.9 & 2\\
                    1 & 0.7 & 0.3 & 2 & 0.5 & 1 & 1\\
                    1& 3 & 0.7 &0 & 0.3 & * & *\\
                    2 & 0.5 & 1 & *&*&*&*
    \end{pmatrix}.
\end{equation}
We then can conduct imputation on $\boldsymbol{z}^*$, which has much fewer missing entries than $\boldsymbol{x}$.
\subsection{Theoretical analysis} 
In this section, we analyze the explained variance of applying PCA on each block compared to PCA on data $\boldsymbol{x}$. Assume we have a data set $\boldsymbol{x}$ with $k$ blocks $\boldsymbol{x}_1,\boldsymbol{x}_2,...,\boldsymbol{x}_k$ where $\boldsymbol{x}_i$ has the size $p_i \times n$ $(1 \le i \le k)$. Let $\mathbf{S}$ denote the covariance matrix estimation of $\boldsymbol{x}$ and $\mathbf{S}_i$ denote the covariance matrix estimation of block $i^{th}$. Here, $\mathbf{S}_i$ is a principal sub-matrix of $\mathbf{S}$.
Now, considering $i^{th}$-block and suppose that applying PCA on $\boldsymbol{x}_i$ will reduce the dimension from $p_i$ to $q_i(<p_i)$. So the explained variance of applying PCA upon $\boldsymbol{x}_i$ is
\begin{equation}
    \mathbf{EV}^{(i)}_{q_i} = \frac{\sum^{q_i}_{j=1}\lambda^{(i)}_j}{\sum^{p_i}_{j=1}\lambda^{(i)}_j}.100\%
\end{equation}
where $\lambda^{(i)}_j$ is the $j^{th}$ eigenvalue of non-increasing eigenvalues of $\mathbf{S}_i$. i.e. $\lambda^{(i)}_1 \ge \lambda^{(i)}_2\ge\dots\ge\lambda^{(i)}_{p_i}$.\\
Let $q=q_1+q_2+\dots+q_k$. The explained variance of applying PCA on $\boldsymbol{x}$ reduce the dimension from $p$ to $q$ presents
\begin{equation}
    \mathbf{EV}_q=\frac{\sum^{q}_{j=1}\lambda_j}{\sum^{p}_{j=1}\lambda_j}.100\%
\end{equation}
where $\lambda_j$ is the $j^{th}$ eigenvalue of non-increasing eigenvalues of $\mathbf{S}$. The lower bound and upper bound of the explained variance mean of all blocks are provided
\begin{equation}\label{eq lower-upper}
    k.\frac{\lambda_{p-\min{(p_{i}-q_{i})}}}{\sum^{p}_{j=1}\lambda_j} \le \frac{1}{k}\sum^k_{i=1}\textbf{EV}^{(i)}_{q_i} \le 1 - k.\frac{\lambda_p}{\sum^{p}_{j=1}\lambda_j}
\end{equation}

\textit{Proof.} For each block $\boldsymbol{x}_i$, note that $\mathbf{S}$ is Hermitian and positive semidefinite matrix, and $\mathbf{S}_i$ is a principal sub-matrix of $\mathbf{S}$, by the well known Cauchy’s interlacing theorem \cite{gowda2011cauchy} we have
\begin{align}
    \lambda_j &\ge \lambda_j^{(i)} \ge \lambda_{j+(p-p_i)}, \quad &j=1,2,\dots,p_i. \\
    \lambda_l &\ge 0,  &l=1,2,\dots,p.
\end{align}
Then,
\begin{equation}
    \sum^{q_i}_{j=1}\lambda_j^{(i)} \ge \sum^{q_i}_{j=1}\lambda_{j+p-p_i} \ge \lambda_{q_i+p-p_i}, \quad i = 1,2,\dots,k.
\end{equation}
Therefore, for all $i \in \{1,2,\dots,k \}$:
\begin{equation}
    \sum^{q_i}_{j=1}\lambda_j^{(i)} \ge \lambda_{\max(q_i+p-p_i)} = \lambda_{p-\min(p_i-q_i)}
\end{equation}
This implies that
\begin{align}
    \frac{1}{k}\sum^k_{i=1}\textbf{EV}^{(i)}_{q_i} &= \frac{1}{k}\sum^k_{i=1}\frac{\sum^{q_i}_{j=1}\lambda^{(i)}_j}{\sum^{p_i}_{j=1}\lambda^{(i)}_j} \nonumber\\
    &\ge \frac{\lambda_{p-\min(p_i-q_i)}}{k}\sum^k_{i=1}\frac{1}{\sum^{p_i}_{j=1}\lambda^{(i)}_j} \\
    &\ge \frac{\lambda_{p-\min(p_i-q_i)}}{k}.\frac{k^2}{\sum^k_{i=1}\sum^{p_i}_{j=1}\lambda^{(i)}_j} \label{eq lower}
\end{align}
On the other hand, $\mathbf{S}_1,\mathbf{S}_2,\dots,\mathbf{S}_k$ are partition of $\mathbf{S}$ so
\begin{equation}\label{eq trace}
    Tr(\mathbf{S}_1)+Tr(\mathbf{S}_2)+\dots+Tr(\mathbf{S}_k) = Tr(\mathbf{S})
\end{equation}
where $Tr(\mathbf{S})$ is the sum of elements on the main diagonal of $\mathbf{S}$. By the property of eigenvalue, 
\begin{equation}\label{eq property eigen}
    \sum^{p_i}_{j=1}\lambda_j^{(i)} = Tr(\mathbf{S}_i), \qquad i = 1,2,\dots,k.
\end{equation}
From \eqref{eq trace} and \eqref{eq property eigen} then 
\begin{equation}\label{eq eq}
    \sum^k_{i=1}\sum^{p_i}_{j=1}\lambda^{(i)}_j = \sum^k_{i=1}Tr(\mathbf{S}_i) = Tr(\mathbf{S}) = \sum^{p}_{j=1}\lambda_j
\end{equation}
Thus, the inequality \eqref{eq lower} becomes
\begin{equation}
    \frac{1}{k}\sum^k_{i=1}\textbf{EV}^{(i)}_{q_i} \ge k.\frac{\lambda_{p-\min(p_i-q_i)}}{\sum^{p}_{j=1}\lambda_j}
\end{equation}
For the upper bound, 
\begin{align}
     \frac{1}{k}\sum^k_{i=1}\textbf{EV}^{(i)}_{q_i} &= \frac{1}{k}\sum^k_{i=1}\frac{\sum^{q_i}_{j=1}\lambda^{(i)}_j}{\sum^{p_i}_{j=1}\lambda^{(i)}_j}\\
     & = \frac{1}{k}\sum^k_{i=1}\left(1 - \frac{\sum^{p_i}_{j=q_i+1}\lambda^{(i)}_j}{\sum^{p_i}_{j=1}\lambda^{(i)}_j} \right)\\
     & = 1 - \frac{1}{k}\sum^k_{i=1}\frac{\sum^{p_i}_{j=q_i+1}\lambda^{(i)}_j}{\sum^{p_i}_{j=1}\lambda^{(i)}_j} \label{eq1 upper}
\end{align}
Furthermore, for all $i \in \{1,2,\dots,k\}$ we get
\begin{equation}\label{eq2 upper}
    \sum^{p_i}_{j=q_i+1}\lambda_j^{(i)} \ge \sum^{p_i}_{j=q_i+1}\lambda_{j+(p-p_i)} \ge \lambda_{p}
\end{equation}
Together, \eqref{eq1 upper} and \eqref{eq2 upper} imply that
\begin{align}
     \frac{1}{k}\sum^k_{i=1}\textbf{EV}^{(i)}_{q_i} & \le 1 - \frac{\lambda_p}{k}\sum^k_{i=1}\frac{1}{\sum^{p_i}_{j=1}\lambda^{(i)}_j}\\
     & \le 1 - \frac{\lambda_p}{k}.\frac{k^2}{\sum^k_{i=1}\sum^{p_i}_{j=1}\lambda^{(i)}_j}
\end{align}
By \eqref{eq eq} then
\begin{equation}
    \frac{1}{k}\sum^k_{i=1}\textbf{EV}^{(i)}_{q_i} \le 1 - k.\frac{\lambda_p}{\sum^p_{j=1}\lambda_j}
\end{equation}

The inequality \eqref{eq lower-upper} shows that the upper bound and lower bound depend on the number of blocks of data $\boldsymbol{x}$. When $k$ is higher, the upper bound of the explained variance mean will be smaller. Therefore, we should consider applying the strategy of PCA upon each block of data if the upper bound is smaller than the expectation of explained variance.
 

\section{Experiments}\label{experiment}

\subsection{Experimental setting}



To illustrate that BPI can work with various imputation techniques, we employed GAIN \cite{yoon2018gain}, and Soft Impute \cite{mazumder2010spectral} as imputation methods. The experiments are conducted on three datasets: MNIST \cite{lecun1998MNIST}, Fashion MNIST \cite{xiao2017fashion}, each of 10 classes, 784 features, and 70000 samples. In addition, we also use the RNA-Seq (HiSeq) PANCAN dataset \cite{misc_gene_expression_cancer_rna-seq_401}, a dataset of 5 classes, where the number of features is significantly higher than the number of samples (20531 features versus 801 samples). For each dataset, we emulate the monotone missing pattern. For example, on the MNIST dataset, we remove the pixels at the bottom right corner of each image. We selected 6000 samples for each Fashion MNIST, MNIST, and the entire PANCAN dataset. We emulated the monotone missing condition for Fashion MNIST and MNIST datasets by dividing the dataset into four partitions and removing 100, 200, and 300 features for the second and third partitions, respectively. For the PANCAN dataset, the number of missing features for the second, third, and fourth partitions is 2000, 4000, and 6000, respectively. A summary of the datasets and the number of missing features can be found in table \ref{table_info_datasets}.


All experiments were done on a computer with 16GB of RAM and 6 CPU cores, running the Linux operating system on an Intel I5-9400F, 4.100 Ghz. The link to the source codes will be provided upon the acceptance of the paper. For the results, we refer to the \textbf{baseline} when the data is imputed directly with an imputation method, and then PCA is applied to the imputed data, and finally, a classifier is trained on the resulting data. Meanwhile, \textbf{BPI} refers to imputing data based on the BPI framework along with an imputation technique and then training a classifier on the resulting dataset.    

\begin{table*}[!h]
\caption{Datasets under study. The number of missing features is a tuple of 3, the first, second, and third element being the number of missing features of the datasets' second, third, and fourth partitions, respectively.}
\label{table_info_datasets}
\centering
\begin{tabular}{ccccc}
\hline
Dataset                                               & \# Classes & \# Features & \# Samples & \# Missing feature   \\ \hline
MNIST \cite{lecun1998MNIST}          & 10         & 784         & 6000       & (100, 200, 300)    \\ \hline
Fashion MNIST \cite{xiao2017fashion} & 10         & 784         & 6000       & (100, 200, 300)    \\ \hline
 PANCAN 
 \cite{misc_gene_expression_cancer_rna-seq_401} & 5          & 20531       & 801        & (2000, 4000, 6000) \\ \hline
\end{tabular}
\end{table*}


\subsection{Result \& Analysis}

\begin{figure*}[!h]
\centering
    \includegraphics[trim=0cm 0cm 0cm 0cm, clip,width=12cm]{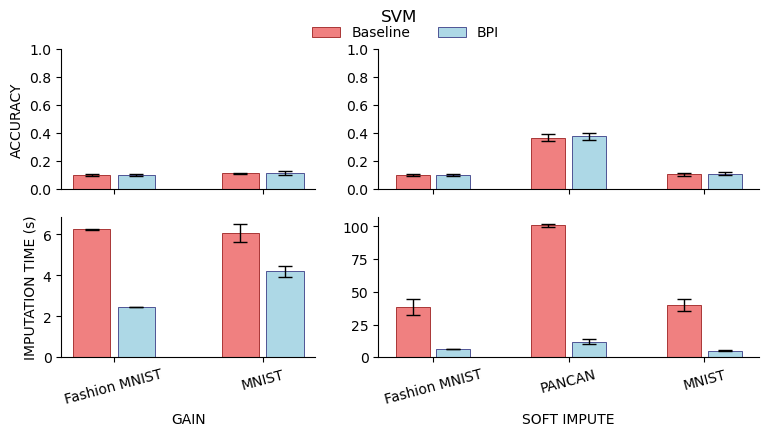}
    \caption{Time and accuracy comparison between \textbf{Baseline} and BPI with different imputation methods across various datasets using SVM classifier.}
    \label{fig:svm_result}
\end{figure*}

\begin{figure*}[!h]
\centering
    \includegraphics[trim=0cm 0cm 0cm 0cm,clip,width=12cm]{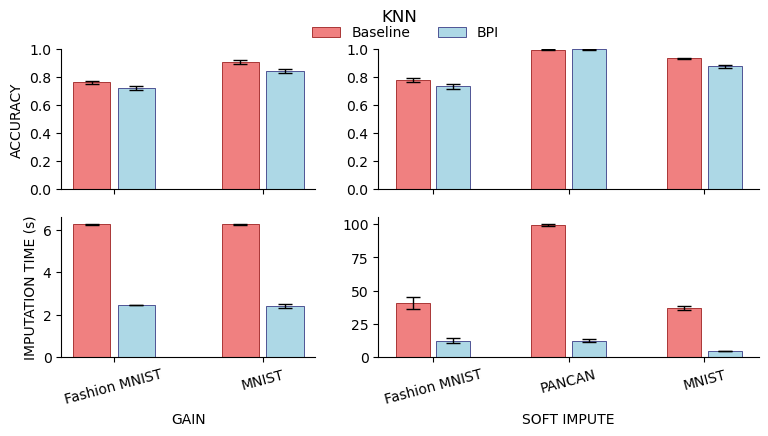}
    \caption{Time and accuracy comparison between \textbf{Baseline} and BPI with different imputation methods across various datasets using KNN classifier.}
    \label{fig:knn_result}
\end{figure*}

\begin{figure*}[!h]
\centering
    \includegraphics[trim=0cm 0cm 0cm 0cm,clip,width=12cm]{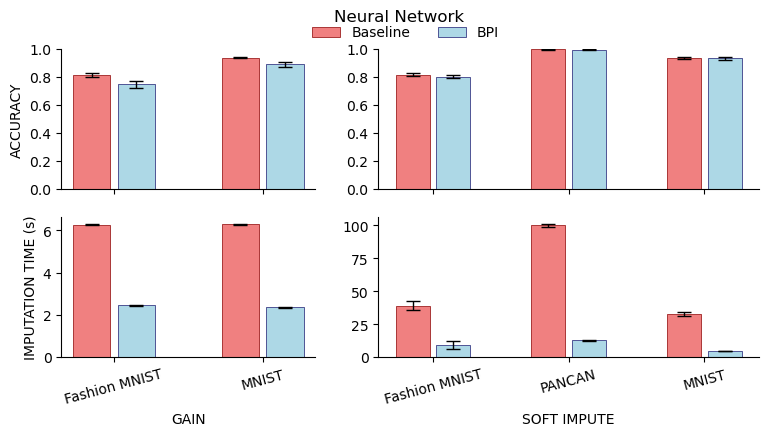}
    \caption{Time and accuracy comparison between \textbf{Baseline} and BPI with different imputation methods across various datasets using a Neural Network classifier.}
    \label{fig:nn_result}
\end{figure*}


Across all datasets, we observed significant reductions in the imputation time of BPI compared to the corresponding baseline, ranging from 52\% to 88\%. In particular, reductions for the Fashion MNIST datasets are 61\% and 76\% for GAIN and SOFT IMPUTE methods, respectively. For the MNIST dataset, the reductions are 52\% and 87\% for the two methods. Notably, for the RNA-Seq (HiSeq) PANCAN dataset, we observed an 88\% reduction in imputation time for SOFT IMPUTE. A common trend among the three datasets is that we observed the most significant reduction in imputation time with Soft Impute and the least significant reduction in the GAIN method. Notably, while imputing the RNA-Seq (HiSeq)
PANCAN dataset with baseline Soft Impute took 100 seconds; it only took 12 seconds for BPI Soft Impute.

Interestingly, with the same imputation method on the RNA-Seq (HiSeq) PANCAN dataset, we see an improvement of 0.3\% and 2.8\% for BPI compared to the corresponding baseline on the KNN and SVM classifier and a slight decrease of 0.3\% on the Neural Network Classifier. Note that this is a dataset where the number of features is significantly higher than the number of observations. 
On the other hand, for the GAIN imputation on the Fashion MNIST dataset, we observed a decrease of 5.4\%, 8.1\%, and 4.5\%  of the BPI method compared to the corresponding baseline for the KNN, Neural Network, and SVM classifier. With the same imputation method on the MNIST dataset, we observed a 7\% and 5\% decrease for the KNN and Neural Network classifier and a 1.7\% improvement in the SVM classifier. For the Soft Impute method on the Fashion MNIST dataset, we see a reduction of 6.3\%, 1.86\%, and 1.87\% in accuracy for the KNN, Neural Network, and SVM classifier, respectively. 

For the MNIST dataset, we see a drop of 6.2\% and 0.1\% for the KNN and Neural Network classifiers and an improvement of 2.6\% on the SVM classifier. 
In summary, we see a slight compromise in accuracy trade-off for a significant reduction in imputation time for the BPI framework. More details of the results are reported in table \ref{tab:result-detail}.



\begin{table*}[!h]
\renewcommand{\arraystretch}{1.5}
\caption{A summary of classification accuracy and imputation time comparison between the Baseline and BPI across different datasets, classifiers, and imputation methods. Each experiment was repeated 10 times to calculate the mean and standard deviation for accuracy and imputation time. For the the RNA-Seq (HiSeq) PANCAN  dataset, since the number of features is higher than the number of samples, we did not perform GAIN  in the baseline method.}
\label{tab:result-detail}
\centering
\begin{tabular}{@{}ccc|cc|cc@{}}

\hline
\multirow{2}{*}{\textbf{Imputer}} 
&\multirow{2}{*}{\textbf{Dataset}} 
&\multirow{2}{*}{\textbf{Classifier}} 
&\multicolumn{2}{c|}{\textbf{Accuracy}} 
&\multicolumn{2}{c}{\textbf{Imputation time (second)}} \\ 
\cline{4-7} 
&&& \textbf{Baseline} & \textbf{BPI} & \textbf{Baseline} & \textbf{BPI}
  \\ \hline
\multirow{9}{*}{GAIN} &
  \multirow{3}{*}{Fashion MNIST} &
  KNN &
  \textbf{0.762 ± 0.009} &
  0.721 ± 0.015 & 
  6.253 ± 0.016 & 
  \textbf{2.444 ± 0.011} \\ \cline{3-7} 
 &
   &
  Neural Network &
  \textbf{0.814 ± 0.014} &
  0.748 ± 0.025 &
  6.261 ± 0.030 &
  \textbf{2.444 ± 0.020} \\ \cline{3-7} 
 &
   &
  SVM &
  \textbf{0.104 ± 0.007} &
  0.100 ± 0.007 &
  6.244 ± 0.034 &
  \textbf{2.450 ± 0.017} \\ \cline{2-7} 
 &
  \multirow{3}{*}{\begin{tabular}[c]{@{}l@{}} PANCAN\end{tabular}} &
  KNN & NA
   &
  0.995 ± 0.005 & NA
   &
  11.938 ± 0.496 \\ \cline{3-7} 
 &
   &
  Neural Network &
  NA &
  0.946 ± 0.045 & NA
   &
  11.850 ± 0.570 \\ \cline{3-7} 
 &
   &
  SVM &
   NA&
  0.368 ± 0.023 &
  NA &
  12.310 ± 0.651 \\ \cline{2-7} 
 &
  \multirow{3}{*}{MNIST} &
  KNN &
  \textbf{0.908 ± 0.011} &
  0.844 ± 0.013 &
  6.245 ± 0.020 &
  \textbf{2.400 ± 0.099} \\ \cline{3-7} 
 &
   &
  Neural Network &
  \textbf{0.939 ± 0.005} &
  0.891 ± 0.015 &
  6.279 ± 0.022 &
  \textbf{2.363 ± 0.018} \\ \cline{3-7} 
 &
   &
  SVM &
  0.114 ± 0.003 &
  \textbf{0.116 ± 0.012} &
  6.083 ± 0.411 &
  \textbf{4.200 ± 0.244} \\ \hline
\multirow{9}{*}{Soft Impute} &
  \multirow{3}{*}{Fashion MNIST} &
  KNN &
  \textbf{0.783 ± 0.013} &
  0.734 ± 0.017 &
  40.939 ± 4.288 &
  12.346 ± 1.859 \\ \cline{3-7} 
 &
   &
  Neural Network &
  \textbf{0.818 ± 0.009} &
  0.803 ± 0.010 &
  39.040 ± 3.387 &
  9.278 ± 2.820 \\ \cline{3-7} 
 &
   &
  SVM &
  \textbf{0.102 ± 0.009} &
  0.100 ± 0.006 &
  38.328 ± 5.663 &
  6.223 ± 0.103 \\ \cline{2-7} 
 &
  \multirow{3}{*}{\begin{tabular}[c]{@{}l@{}}  PANCAN\end{tabular}} &
  KNN &
  0.995 ± 0.004 &
  \textbf{0.998 ± 0.003} &
  99.574 ± 0.629 &
  \textbf{12.293 ± 1.095} \\ \cline{3-7} 
 &
   &
  Neural Network &
  \textbf{0.999 ± 0.002} &
  0.996 ± 0.004 &
  100.010 ± 0.964 &
  \textbf{12.793 ± 0.411} \\ \cline{3-7} 
 &
   &
  SVM &
  0.369 ± 0.026 &
  \textbf{0.380 ± 0.024} &
  100.990 ± 0.967 &
  \textbf{11.894 ± 1.692} \\ \cline{2-7} 
 &
  \multirow{3}{*}{MNIST} &
  KNN &
  \textbf{0.934 ± 0.004} &
  0.877 ± 0.009 &
  36.716 ± 1.352 &
  \textbf{4.759 ± 0.061} \\ \cline{3-7} 
 &
   &
  Neural Network &
  \textbf{0.936 ± 0.008} &
  0.934 ± 0.009 &
  32.907 ± 1.361 &
  \textbf{4.799 ± 0.097} \\ \cline{3-7} 
 &
   &
  SVM &
  0.108 ± 0.010 &
  \textbf{0.110 ± 0.011} &
  40.181 ± 4.401 &
  \textbf{5.001 ± 0.337} \\ \hline
\end{tabular}

\end{table*}


\section{Conclusion}\label{conclusion}

In this paper, we introduced BPI, a novel dimensionality reduction-imputation framework that can speed up the running time significantly compared to using PCA after imputing the missing data.   
This efficiency gain is particularly important in scenarios where time constraints are a crucial factor, especially for high-dimensional datasets.  
However, it is important to note that our proposed technique has certain limitations. Specifically, a drawback of the proposed technique is that PCA requires continuous data. For categorical data, a potential solution is to use one-hot encoding or learned embedding, and we will perform BPI for categorical data in the future. 
Additionally, since PCA has the denoising property, it may improve the imputation quality or classification accuracy for noisy datasets. In the future, we would like to investigate this through noisy datasets.





\bibliographystyle{IEEEtran}
\bibliography{ref}

\end{document}